\pdfoutput=1
\documentclass[11pt]{article}

\usepackage[final]{acl}

\usepackage{times}
\usepackage{latexsym}

\usepackage[T1]{fontenc}
\usepackage[utf8]{inputenc}

\usepackage{microtype}

\usepackage{inconsolata}

\usepackage{graphicx}
\usepackage{graphicx}
\usepackage{booktabs}
\usepackage[normalem]{ulem}
\usepackage{algorithm}
\usepackage{algpseudocode}
\usepackage{amsmath}
\usepackage{multirow} 
\usepackage{makecell}

\useunder{\uline}{\ul}{}

\usepackage{xcolor}
\newcommand{\jose}[1]{\textcolor{black}{#1}}
\newcommand{\dimos}[1]{\textcolor{black}{#1}}
\newcommand{\fr}[1]{\textcolor{black}{#1}}
\newcommand\DATASET{X-Sensitive} % dataset name work in progress
\makeatletter
\newcommand\footnoteref[1]{\protected@xdef\@thefnmark{\ref{#1}}\@footnotemark}
\makeatother

\newcommand\blfootnote[1]{%
  \begingroup
  \renewcommand\thefootnote{}\footnote{#1}%
  \addtocounter{footnote}{-1}%
  \endgroup
}

\title{Sensitive Content Classification in Social Media: \\ A Holistic Resource and Evaluation}

\author{Dimosthenis Antypas\textsuperscript{1*} Indira Sen\textsuperscript{2*$\dagger$}  Carla Perez-Almendros\textsuperscript{1}\\ \textbf{Jose Camacho Collados\textsuperscript{1} Francesco Barbieri\textsuperscript{$\dagger$}}\vspace{0.1cm}
\\$^1$Cardiff NLP, School of Computer Science and Informatics, Cardiff University, UK
\\$^2$University of Mannheim, Germany \vspace{0.1cm}
\\$^1$\texttt{\{AntypasD,PerezAlmendrosC,CamachoColladosJ\}@cardiff.ac.uk} 
\\$^2$\texttt{indira.sen@uni-mannheim.de}
}

\begin{document}
\maketitle
\begin{abstract}
The detection of sensitive content in large datasets is crucial for ensuring that shared and analysed data is free from harmful material. However, current moderation tools, such as external APIs, suffer from limitations in customisation, accuracy across diverse sensitive categories, and privacy concerns. Additionally, existing datasets and open-source models focus predominantly on toxic language, leaving gaps in detecting other sensitive categories such as substance abuse or self-harm. In this paper, we put forward a unified dataset tailored for social media content moderation across six sensitive categories: conflictual language, profanity, sexually explicit material, drug-related content, self-harm, and spam. By collecting and annotating data with consistent retrieval strategies and guidelines, we address the shortcomings of previous focalised research. 
Our analysis demonstrates that fine-tuning large language models (LLMs) on this novel dataset yields significant improvements in detection performance compared to open off-the-shelf models such as LLaMA, and even proprietary OpenAI models, which underperform by 10-15\% overall. This limitation is even more pronounced on popular moderation APIs, which cannot be easily tailored to specific sensitive content categories, among others. 
\blfootnote{$^{*}$Equal contribution.}
\blfootnote{$\dagger$ Work done while at Snap Inc.}
\end{abstract}

\section*{Disclaimer}
Due to the nature of the subject studied in this work, \textbf{this paper contains sensitive and potentially offensive language. Reader discretion is advised.}

\section{Introduction}

Consider the case of a researcher or a data analyst who needs to filter sensitive content from a large dataset. Such task is crucial to ensure that data shared or analysed does not include harmful or inappropriate material. One might initially consider using external tools like Perspective\footnote{\url{https://perspectiveapi.com/}} or OpenAI moderation APIs\footnote{\url{https://platform.openai.com/docs/guides/moderation}} to assess and filter sensitive content. However, this approach often falls short, presenting important limitations for an effective identification of inappropriate content online \cite{udupa2023ethical}. For instance, they usually offer limited customisation capabilities (e.g., how can the model be improved if it fails on specific domains or keywords?), and limited sensitive categories coverage (lacking in detecting self-harm \cite{uban2020deep}, for example).
%they are trained on \textit{popular} types of sensitive language, such as abusive and offensive language or hate speech,  neglecting more subtle but equally important types of sensitive content, such as self-harm discourses . 
Finally, these tools rely on external servers, which raises concerns about data privacy and security \cite{oseni2021security, gupta2023chatgpt}.

Alternatively, one might consider using existing datasets and open-source models for sensitive content detection. This could be a viable option if the primary focus was on detecting toxic language, given the abundance of resources available in this area. However, if the goal extends to identifying additional sensitive categories such as sexually explicit content, drugs, self-harm or spam, the situation becomes more challenging. Data on these less-explored categories is limited and sometimes outdated. %\fr{\textit{we can remove this paragraph if we don\'t have those references}} 
For instance, those categories could be covered on datasets that are biased \cite{wiegand2019detection}, old or inaccessible even in an anonymized manner~\cite{tadesse2019detection,sawhney2018exploring}, too small-scaled or rely on an handful of keywords to extract the data~\cite{ding2016analyzing}. This limited approach can result in incomplete or less accurate detection of sensitive content. 

Existing solutions either require sending data to external servers or fail to address the full spectrum of sensitive content categories. %Furthermore, these resources do not provide the flexibility for users to customize models according to their specific needs.
In response to these challenges, this paper proposes a \jose{new holistic approach}: \textbf{a unified dataset for detecting sensitive text across a broad range of categories}, including (1) conflictual language, (2) profanity, (3) sexually explicit material, (4) drug-related content, (5) self-harm, and (6) spam\footnote{We use social media platform user guidelines to be sure to have reliable guidelines.}. \jose{This dataset can then be used for both evaluation and fine-tuned models to all these categories under a single framework.}

Our approach involves collecting and re-annotating data to ensure consistent quality across sensitive classes. %to ensure consistent data distribution (same retrieval strategies, topics, source platforms), and annotations guidelines and quality.
The alternative of \jose{putting} together a collection of existing datasets to create one single dataset would include several limitations, as (1) there would be bias towards the data distribution (different retrieval strategies, topics, source platforms), (2) annotation guidelines and quality would differ, and (3) each text would include only one sensitive dimension (even if the text includes multiple sensitive categories).

In short, we propose a holistic approach when it comes to sensitive content moderation in social media, overcoming common shortcomings of previous works and providing the following contributions:
\begin{itemize}
\item \jose{New dataset:} We introduce the X-Sensitive dataset, manually annotated and tailored for social media content, featuring multiple categories and designed to be resilient against keyword and domain shifts.
\item %Study the interplay between sensitive categories and as well how these categories vary for different annotators demographics.
\jose{Sensitive category analysis: We} study the interplay between sensitive categories and how these categories vary across different annotator demographics.
\item \jose{Model evaluation}: The best results are achieved from large language models (LLMs) with 8 billion parameters, fine-tuned on our dataset. However, smaller language models\footnote{Pre-trained on social media language.} (355 million parameters)
show only about 2\% less accuracy.
\item Comparison with off-the-shelf LLMs: We find that readily available LLMs, such as 
%chatGPT-turbo
gpt-4o, under-perform by 10-15\% compared to fine-tuned models, highlighting the value of bespoke training on specialised datasets.
\end{itemize}

%\fr{TODO: spans/highlights? Appendinx / future work}

% BULLET POINT OF FINDING/PAPER NOVELTIES:
% \begin{itemize}
    % \item New sensitive content dataset for social media content, that include multiple classes and as well is robust to keyword/domain shift
    % \item Sensitive content API have big limitations: 1. users do not have control on how the sensitive content is trained, 2. they lack in detecting some sensitive categories
    % \item best models are LLMs (70B) fine-tuned on our dataset, but smaller (355M) LM pre-trained on social media language do not fall far behind (about 2\% less in accuracy)
    % \item Out of the shelf LLMs (e.g. chatGPT-turbo) perform 10-15\% worse than fine-tuned models
% \end{itemize}

%----We did this?----
%\textbf{Annotator demographics}
%We also have annotator demographics, so we can compare agreement and disagreement by different backgrounds
%---------------------

The {\DATASET} dataset, as well as the best performing models built upon it, are made openly available. {\DATASET} is available at \url{https://huggingface.co/datasets/cardiffnlp/x_sensitive}. Best multi-label and binary  models are available at \url{https://huggingface.co/cardiffnlp/twitter-roberta-large-} \url{sensitive-multilabel} and \url{https://huggingface.co/cardiffnlp/twitter-roberta-large-sensitive-binary}, respectively.

\section{Related Work}

Our current work aims to bridge the gap between current academic research in content moderation and the needs of content moderators in realistic scenarios. While hate speech and toxic language are widely studied in NLP, there is little research on other types of sensitive content that platforms seek to detect and moderate, such as sexually explicit content or content about illicit substances~\cite{arora2023detecting}. To that end, our work is situated at the intersection of NLP research on harmful language detection and research on platform governance and content moderation.

\subsection{Automatic Detection of Harmful Language}

\paragraph{Hate speech Detection.} Automatic detection of hate speech, and related social constructs like offensive and toxic language, is an active area of research in NLP~\cite{fortuna2018survey,poletto2021resources}. However, there are several challenges, not least the lack of high quality datasets for studying such phenomena~\cite{vidgen2020directions}. 

\paragraph{Self-harm and Suicidal Content Detection. }\citet{chancellor2016quantifying} identify communities with self-harm related content, while~\citet{tejaswini2024depression} also look into related behaviors such as depression. Previous research has also looked into suicidal content detection~\cite{coppersmith2018natural} and general self-harm~\cite{un2021towards}. There are generally several ethical challenges associated with studying mental health conditions, including self-harm and suicidal ideation~\cite{chancellor2019taxonomy}.

\paragraph{Illicit Substance Abuse. }Past research has looked into automated approaches for discussions of illegal or banned substances, including drugs~\cite{buntain2015your,lavanya2022auto,simpson2018detecting}.

% \textbf{Sexually Explicit Content. }Past research has also developed automated approaches to detection sexually explicit content~\cite{barrientos2020machine}, sexual harassment~\cite{chowdhury2019speak}, and sexualized cyberbullying~\cite{basu2021cyberpolice}.

\paragraph{Sexually Explicit Content. } Research has also focused on developing automated systems to detect sexually explicit content~\cite{barrientos2020machine}, address sexual harassment~\cite{chowdhury2019speak}, and identify sexualised cyberbullying~\cite{basu2021cyberpolice}.

\paragraph{Spam Detection. }Automatic Spam detection is widely studied in NLP as well as computer security communities. Typical automation techniques rely on expert-annotated training data used to train machine learning models~\cite{hussain2019spam}. However, like the other categories spam detection has rarely been studied in the context of other types of problematic content, with~\citet{founta2018large} being an exception.  

\subsection{Content Moderation and Platform Governance} 

Platforms on the internet, such as web and social media sites, often employ mechanisms to curate their content and reduce problematic or harmful content through content moderation (CM). CM can take many forms, from commercial content moderation outsources to underpaid moderators in the Global South~\cite{roberts2019behind} to artisanal solutions, some of which are led by volunteers~\cite{caplan2018content}. Yet as content grows, platforms turn towards automated methods, often Artificial Intelligence (AI) based techniques either solve or ameliorate their moderation problem~\cite{gorwa2020algorithmic}.

However, the question remains on how much of this detection is automatable?~\cite{gillespie2020content}. There are not only several technological limitations (e.g. the dearth of AI methods for non-English content~\cite{vidgen2020directions}) but also political challenges (e.g., who gets to decide what is harmful?~\cite{fleisig2024perspectivist}) and challenges at the nexus of technology and politics (e.g., how do we aggregate the potentially divergent judgements of whether something is harmful?~\cite{fan2020digital,gordon2022jury}). On the other hand, platform studies researchers have studied which types of technological solutions, including AI-based tools, would facilitate the work of content moderators while also establishing some of the tensions of the whole practice of content moderation. However, it is unclear if those proposing technological solutions for CM are basing their solutions on the requirements of content moderators. 

\textbf{Categories of Sensitive Content. }Several researchers have attempted to categorise what counts as `sensitive' content on web and social media platforms, i.e., content that requires moderation \cite{jiang2020characterizing,scheuerman2021framework}. 

We address one of the many challenges of automatic content moderation --- lack of benchmark datasets for measuring understudied categories of problematic content like discussion related to self-harm and illicit substances, particularly drugs. We also provide a holistic benchmark of both these aforementioned understudied categories as well as widely studied categories like profanity, allowing researchers to model the associations between different types of sensitive content.

\section{ {\DATASET} Dataset}

In order to study sensitive content in X, we construct a new dataset, {\DATASET}. As a first step, we conceptualise a topic taxonomy based on community guidelines from several social media platforms.

\subsection{Taxonomy}
%\textbf{Conceptualising the Taxonomy.}
We use the community guidelines of various social media platforms to ground our taxonomy~\cite{scheuerman2021framework}. Using iterative coding, we refine, merge, and fix 5 broad categories and 7 specific subcategories of sensitive content which are mapped to rules in community guidelines. Our final categories and their definitions are:

% \item Illicit Content
% \begin{enumerate}
%     \item Regulated Goods
%     \begin{enumerate}
% \item \textbf{Weapons:} Content that encourages, promotes or glorifies the use of weapons or firearms. Also applicable to content that mentions sales, purchases, or the act of obtaining or trying to obtain firearms or weapons
\paragraph{Drugs.} Content that encourages, promotes or glorifies the use of regulated drugs. Also applicable to content that mentions sales, purchases, or the act of obtaining or trying to obtain regulated drugs.
        
\paragraph{Sexually Explicit Content (Sex).} Pornographic or other types of sexual content.  We collect and download 50+ textual abusive language datasets from hatespeechdata.com. We then use the Perspective API to label these datasets with the ‘sexually explicit’ endpoint and then use the labelled data for fine-tuning the XLM-T sexually explicit content classifier.

\paragraph{Hate speech.} Attacks against protected attributes like race, colour, caste, ethnicity, national origin, religion, sexual orientation, gender identity, disability, or veteran status, immigration status, socio-economic status, age, weight or pregnancy status.

\paragraph{Other conflictual language.} Attacks based on other categories or without any mention of the categories mentioned.
% TODO
% We use the textual dataset obtained from hatespeechdata.com for training the conflictual language classifiers. Most of these datasets model adversarial behaviour like racism, sexism, hate speech, and others. These dataset are typically curated from social media data with Twitter being a primary source, and labeled manually. However, due to the wide variety of harmful communication, these datasets do not share a common codebook. We differentiate between conflictual and non-conflictual based on the label mapping provided by Ritcsh et al []. Note that due to differences in data sampling and annotation approaches, these datasets might not have a consistent notion of conflictual language. Hence, we consider the classifier fine tuned on this dataset to be a ‘weak’ classifier that provides a holistic, but imprecise notion of conflictual language. Furthermore, while our taxonomy differentiates between hate speech and other conflictual language that is either untargeted or does not target protected attributes, the datasets used for the phase 1 classifiers do not always make this distinction. Indeed, previous research found that human labelers find it difficult to differentiate between hate speech and other types of conflictual language. Hence, this current dataset only allows us to differentiate between conflictual versus non-conflictual language, with the former encapsulating hate speech and other conflictual content.

\paragraph{Profanity.} Language containing slurs and profanity even if they are not directed towards a specific entity.		
% We repeat the same process as we used for the sexually explicit classifier by leveraging the output of the Perspective API. However, in this case we use the ‘profanity’ endpoint instead of the ‘sexually explicit’ one. 

\paragraph{Self-harm.} Posts depicting, promoting or glorifying violence or harm against oneself, such as eating disorders or suicide.
% We collect data from online communities that discuss self-harm related topics like eating disorders and suicidal ideation. For the negative class, we use subreddit covering a wide variety of topics. Specifically we collect the latest posts from subreddits that have at least 50K subscribers. 

\paragraph{Spam.} Irrelevant content that is unsolicited; or content that aims to drive traffic or attention from a conversation on the platform to entities outside the platform.

% We follow a similar procedure for training the spam classifier, as we did for the profanity and sexually explicit content classifier, using the Perspective API’s ‘spam’ endpoint. 

\subsection{Message Collection}
\label{sec:tweet-collection}

Typically previous work on sensitive content detection, particularly hate speech detection, uses a small set of keywords to collect data, which may lead to limited coverage of the resultant datasets~\cite{ousidhoum2021probing}. To tackle this problem, we utilise a keyword expansion technique combining word embeddings \cite{mikolov2013efficient},
trained on tweets~\cite{pennington2014glove}, for keyword expansion and clustering for controlling the expanded sets. The specific algorithm is described as follows:

% Keyword list expansion technique using word2vec:

%     1. Start with a seed list {w1, w2, …} 
    
%     2. Cluster keywords’ W2V vectors into k clusters 
    
%     3. Check and select the relevant clusters
    
%     4.  Get the dot product of the entire unit W2V with the cluster mean vectors
    
%     5. Find n1 words that are closest to the mean of each cluster
    
%     6. For each of these new words find the closest n2 words

\begin{algorithm}
\caption{Keyword List Expansion Technique using Word Embeddings }
\begin{algorithmic}[1]
    \State \textbf{Input:} Seed list $\{w_1, w_2, \ldots\}$
    \State \textbf{Output:} Expanded keyword list
    \State Start with a seed list $\{w_1, w_2, \ldots\}$
    \State Cluster keywords' vectors into $k$ clusters
    \State Check and select the relevant clusters
    \State Compute the dot product $\mathbf{v}_i \cdot \boldsymbol{\mu}_c$ for each word vector $\mathbf{v}_i$ and each cluster mean vector $\boldsymbol{\mu}_c$
    \State Find $n_1$ words that are closest to the mean of each cluster
    \For {each new word}
        \State Find the closest $n_2$ words
    \EndFor
\end{algorithmic}
\end{algorithm}

Due to the variety of the categories in our sensitive content category, we look at multiple sources for seed lists (Appendix \ref{sec:appendix-data}, Table \ref{tab:data-sources}). The conceptual similarities between profanity, sexually explicit content, and conflictual content
%, especially the overlap in unigrams surfaced by our weak classifiers 
%(Appendix \ref{sec:appendix-data}, Figure \ref{fig:word_shift})
as well as the existence of lists that collect keywords related to these three types of discourse, make us opt for a unified seed word list covering these three categories.

For self-harm and spam, uni-grams are not as informative as they are for other categories. Therefore for the former, we look at past research on eating disorders and suicidal ideation~\cite{chancellor2016quantifying,chancellor2021suicide} and obtain phrases (n-grams with a high TF-IDF score) from the Reddit data used to train the self-harm phase 1 classifier. For spam, we use the dataset from~\citet{founta2018large}, particularly the tweets that were labelled containing spam and obtain n-grams from there using a similar method. We manually assess each of the keywords for all categories and remove low precision words like ‘snow’ for drugs. While snow may refer to cocaine in some contexts, most tweets containing it do not use it in that sense. After this manual inspection, we apply our cluster-based keyword expansion technique. We again manually assess the keywords and include only those that are relevant to the category. The final statistics of our keywords are listed in Appendix \ref{sec:appendix-data}, Table \ref{tab:seeds-stats}. 

\subsection{Annotation}
\label{sec:annotation}
Each entry of the dataset was annotated by at least three coders, where each coder had to answer with \textit{yes}, \textit{no}, or \textit{not sure} if the tweet contained any of the sensitive classes. Specifically, for the case of conflictual language the annotators were asked to select whether the tweet contained hate speech or any other form of conflictual language. \dimos{This approach aimed for a more fine grained classification of conflictual language. However, due to low agreement between coders we opted to merge the categories "Hate Speech" and "Other Conflictual Language" into a single class \textit{Conflictual Language}.}%If the coder selected \textit{yes} to any of the classes then they had to highlight which part of the text lead them to this answer. 

A label was assigned to a tweet if at least one annotator answered \textit{yes} and there was no direct opposition from the rest of the coders (i.e. the rest of the coders answered \textit{yes} or \textit{not sure}). We refrained from utilising a majority rule in order to create a more realistic and challenging dataset while also weighting the recall of potentially sensitive content higher. 

The coders who worked on this task were selected and filtered through the Prolific.co platform based on their fluency in English. 
The annotation was performed through an interface created with qualtrics\textsuperscript{XM}\footnote{The annotation guidelines can be found in Figure \ref{fig:appendix_guidelines}, Appendix \ref{sec:appendix_annotation}.}. The coders were also provided with 15 examples of already annotated tweets to help them better understand the task. Finally, we utilised several filters to ensure a high quality of annotations. First, we included a set of test questions randomly inserted in the task which were used to filter out low quality coders. Additionally, coders that finished the task too quickly or provided low quality answers (for example always selecting the same answer) were excluded.

Overall 523 coders from various demographic backgrounds took part in the annotation process. We assessed the quality of the annotation by utilising Krippendorff's Alpha (\textit{Alpha}) \cite{krippendorff2011computing}. The annotators achieve 0.49 \textit{Alpha} when considering all available classes and 0.56 \textit{Alpha} when considering only the presence of sensitive content or not. The scores are in line or better with previous similar studies on toxic and sensitive content \cite{muralikumar2023human,lima2024toxic}.%\footnote{\label{note2}Conflictual hate and Conflictual other classes were grouped together.}.

It is interesting to note that when looking at subgroups of annotators based on their demographics we observe higher agreement between specific groups, mainly younger (0.51 \textit{Alpha} in multi-label setting for people 39 old and younger) and non-binary people  (0.82 \textit{Alpha} in the binary setting). More detailed results can be found in Appendix \ref{sec:appendix_annotation}, Tables \ref{tab:agreement_age} and \ref{tab:agreement_gender}). 

Looking in more detail on how different demographics annotate examples a trend is noticed where younger coders and non-binary annotators tend to be more sensitive to the content and are more likely to flag a tweet as sensitive (Appendix \ref{sec:appendix_annotation}, Tables \ref{tab:coder_age} and \ref{tab:coder_gender}).

The discrepancies in agreement between groups indicate the inherent difficulty of the task while also providing evidence of a greater coverage of sensitive content within \DATASET.

\subsection{Statistics}

{\DATASET} contains a total of 8,000 tweets all related to sensitive content with 49\% of them labelled as one or more of the six sensitive classes available making it a challenging dataset. On average tweets flagged as sensitive are assigned 1.4 labels with maximum assigned labels to a single tweet being 4.

\begin{figure}
    \centering
    \includegraphics[width=\linewidth]{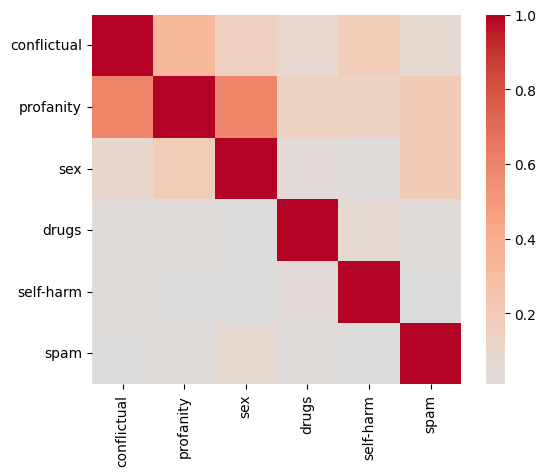}
    \caption{Overlap of classes.}
    \label{fig:class_overlap}
\end{figure}

Our dataset displays a skewed distribution of classes as seen if Table \ref{tab:data-statistics} with \textit{profanity} being the most populated class present (30.4\%). This uneven distribution represents a realistic representation of sensitive content in social media as seen in previous similar studies \cite{beknazar2022toxic} where it estimated a 5\% - 7\% of content displayed is inappropriate, making {\DATASET} ideal for usage in real world applications.

At the same time \textit{profanity} being the most frequent class is also expected. Due to the multi-label nature of the dataset, we expect high overlap between \textit{profanity} and other classes as seen in Figure \ref{fig:class_overlap}. Particularly there is a high overlap between  \textit{profanity} tweets and those labelled as \textit{sexual explicit content}, and \textit{conflictual}.

% \begin{table}[]
% \scalebox{0.85}{
% \begin{tabular}{lccl}
% \toprule
% \textbf{Category} & \textbf{\begin{tabular}[c]{@{}c@{}}Word \\ overlap\end{tabular}} & \textbf{Lev} & \multicolumn{1}{c}{\textbf{Top Highlights}} \\ 
% \toprule
% conflictual & 0.44 & 0.40 & \begin{tabular}[c]{@{}l@{}}nigga bitch racist nigger\\ stupid\end{tabular} \\ 
% \midrule 
% profanity & 0.83 & 0.83 & fucking shit bitch fuck as \\ 
% \midrule
% sex & 0.57 & 0.55 & pussy cock horny dick tit \\ 
% \midrule
% drugs & 0.49 & 0.45 & \begin{tabular}[c]{@{}l@{}}drug pill weed adderall \\ cannabis\end{tabular} \\ 
% \midrule
% self-harm & 0.50 & 0.46 & \begin{tabular}[c]{@{}l@{}}suicide suicidal edtwt \\ thinspo depressed\end{tabular} \\ 
% \midrule
% spam & 0.51 & 0.49 & \begin{tabular}[c]{@{}l@{}}link dm unlock giveaway\\ buy\end{tabular} \\
% \bottomrule
% \end{tabular}
% }
% \caption{Average world overlap and Levenshtein (Lev) distances for the highlights selected by annotators for each category. The top five most frequent terms are also displayed}
% \label{tab:stats_highlights}
%  \end{table}

% \begin{figure}
%     \centering
%     \includegraphics[width=1\linewidth]{images/class_distribution.png}
%     \caption{Distribution of sensitive classes in \DATASET.}
%     \label{fig:class_distribution}
% \end{figure}

In general differences between the classes are revealed even when looking at basic statistics such as the average length of tweets and the presence of emojis in them. As seen in Table \ref{tab:data-statistics} tweets labelled as \textit{spam} tend to be longer on average and include a higher number of emojis, characteristics frequently found on spam messages \cite{robinson2022birds}. Similarly, a higher usage of emojis is observed in tweets flagged as \textit{sexual explicit content}, as specific emojis are often used as representation of sexual acts \cite{thomson2018you}. Furthermore, when examining the top terms of each class based on lexical specificity scores \cite{camacho2016nasari}, a clear distinction between the classes is observed, which also serves as a sanity check for the quality of our dataset.

% As a final quality assurance step, we examine the highlights selected by coders for each label assignment, as shown in Table \ref{tab:stats_highlights}. The most frequently highlighted terms align with our expectations. For instance, the "spam" category commonly includes terms such as "link," "dm," and "giveaway," which are typical in spam messages.

% Furthermore, we assess the agreement among coders by analysing the percentage of word overlap and the Levenshtein distance \cite{levenshtein1966binary}. This analysis reveals that different text segments were important for coders in assigning labels. Notably, in the profanity category, there is a high level of agreement, with a word overlap score and Levenshtein distance score of 0.83. However, for other categories, the agreement is moderate, with average scores of 0.56 for word overlap and 0.53 for Levenshtein distance. These scores highlight the inherent difficulties and subjectivity in this task, as coders may interpret the same text differently.

\section{Experimental Setting}

In this section, we set out the common experimental framework which serve as the basis of the evaluation.

\subsection{Data and Settings}

To evaluate {\DATASET} and establish baselines of its difficulty we establish two distinct settings: binary and multi-label classification. In the binary setting, tweets will be classified into one of two categories,  distinguishing between sensitive and not sensitive content. This approach simplifies the classification process, focusing on the presence or absence of sensitive characteristics in general. In the multi-label setting, tweets can belong to multiple sensitive categories simultaneously, allowing for a more fine-grained analysis that captures the complexity of the content. This dual approach enables a comprehensive evaluation of our dataset's versatility and the classifier's robustness in handling varying degrees of complexity in sensitive content detection.

\begin{table}[]
\scalebox{0.76}{
\centering
\begin{tabular}{lcccl}
\toprule
\textbf{Category} & \textbf{L} & \textbf{Emo} & \multicolumn{1}{c}{\textbf{\%}} & \multicolumn{1}{c}{\textbf{Top Terms}} \\ \toprule
Conflictual & 188.67 & 0.26 & \multicolumn{1}{c}{17.3} & \begin{tabular}[c]{@{}l@{}}fucking racist \\ nigga white shut\end{tabular} \\ \midrule
Profanity & 173.05 & 0.43 & \multicolumn{1}{c}{30.4} & \begin{tabular}[c]{@{}l@{}}fucking shit\\ bitch fuck as\end{tabular} \\ \midrule
Sex & 160.79 & 0.66 & \multicolumn{1}{c}{9.7} & \begin{tabular}[c]{@{}l@{}}cock pussy\\ dick horny cum\end{tabular} \\ \midrule
Drugs & 155.62 & 0.21 & \multicolumn{1}{c}{3.9} & \begin{tabular}[c]{@{}l@{}}drug weed cbd\\ thc mushroom\end{tabular} \\ \hline
Self-harm & 166.35 & 0.44 & \multicolumn{1}{c}{3.0} & \begin{tabular}[c]{@{}l@{}}suicide suicidal\\ attempt commit \\ideation\end{tabular} \\ \midrule
Spam & 200.30 & 1.12 & \multicolumn{1}{c}{3.4} & \begin{tabular}[c]{@{}l@{}}dm project airdrop\\ solana solanaairdrop\end{tabular} \\ \hline
Not Sensitive & 176.39 & 0.32 & \multicolumn{1}{c}{51.2} & \begin{tabular}[c]{@{}l@{}}physically depressing\\ triggering mental \\ depression\end{tabular} \\ 
\midrule
\textbf{Overall} & 174.77 & 0.37 & \multicolumn{2}{c}{} \\
\bottomrule
\end{tabular}}
\caption{General lexical statistics for each class. The averages of the length of tweet, emojis count are reported. The distribution of each class along with the top five terms based on their lexical specificity are also displayed.}
\label{tab:data-statistics}
\end{table}

For both settings we use a split the dataset in train/validation/test sets of 6,000/1,000/2,000 tweets while ensuring that the distribution of classes is similar in each split. To investigate the generalisability capabilities of the models, an additional constraint check is enforced where we ensure that approximately half of the test set, 1,016 tweets, do not share any of the keywords used for collection with tweets from the train set. 

\subsection{Comparison Systems}
\label{sec:comparison-systems}

For the evaluation, we are interested in comparing three types of approaches: fine-tuning on the same dataset (Section \ref{sec:fine}), LLMs with in-context learning either zero- or few-shot (Section \ref{sec:llms}), and out-of-the-box content moderation systems (Section \ref{sec:out-of-the-box}). All the systems are clearly not fully comparable, but our dataset can serve as the basis for establishing this basic ground comparison.

\subsubsection{Fine-tuning}
\label{sec:fine}
We evaluate three distinct models tailored for various applications, including general-purpose and those specialized for social media, each differing in size for our fine-tuning experiments. The large version of \textbf{RoBERTa} \cite{liu2019roberta}  is tested in order to assess the performance of smaller, non specialised, masked language models on our dataset. Time-lm, \textbf{tlm}, \cite{loureiro-etal-2022-timelms}, a RoBERta based model trained on a large X corpus of  154 million tweets is also evaluated to assess the performance of specialised models on social media. The two models are fine-tuned using the implementations provided by Hugging Face \cite{wolf-etal-2020-transformers}  and optimising hyper parameters (learning rate, training epochs, warm-up steps) is conducted using Ray Tune \cite{liaw2018tune}\footnote{Details of the models used can be found in Appendix \ref{sec:appendix-models}.}. Finally, the 8 billion version of Llama-3, \textbf{Llama3-8b}, \cite{llama3modelcard} is also fine-tuned on our dataset by utilising quantisation and PEFT \cite{liu2021p, peft} explore the capabilities of more recent and larger-scale models.

\subsubsection{Zero- and Few-shot}
\label{sec:llms}
In order to assess the zero/few-shot capabilities of large language models in our dataset, we compare four models of different sizes and architectures.

 \noindent\textbf{Llama3}: The 8 and 70 billion instruct versions of Llama3 are tested. These models are designed to follow user instructions more effectively, allowing us to assess how well they adapt in settings where training data is limited or not available. 
 
 \noindent\textbf{chat-gpt-3.5-turbo (chat-gpt):} from OpenaAI, \footnote{\url{https://openai.com/chatgpt}} an encoder/decoder model with approximately 175 billion parameters \cite{brown2020language}. 

 \noindent\textbf{gpt-4o:} the currently latest model from OpenAI which significantly outperforms its predecessor.
 
 Assessing the performance in zero- and few-shot settings, helps us to exlore the capabilities and limitations of these large language models for sensitive content detection.

\begin{table*}[]
\scalebox{0.78}{
\begin{tabular}{ll|cc|ccccccc}
\toprule 
\textbf{Training} & \textbf{Model} & \textbf{Binary} & \textbf{multi-label} & \textbf{Conflictual} & \textbf{Profaninty} & \textbf{Sex} & \textbf{Drugs} & \textbf{Self-harm} & \textbf{Spam} & \textbf{Not sens.} \\ \hline
\multirow{3}{*}{fine-tuned} & RoBERTa & 82.4 & 64.7 & 60.6 & 88.9 & 81.6 & 52.3 & 34.3 & 52.0 & 83.3 \\
 & tlm & 84.4 & 67.7 & 59.6 & 88.8 & 84.3 & 48.9 & 50.6 & 59.1 & 82.4 \\
 & llama3-8b & \textbf{85.6} & \textbf{69.8} & 61.7 & \textbf{90.6} & \textbf{85.8} & 53.9 & 50.6 & 61.2 & \textbf{85.1} \\ \hline
\multirow{3}{*}{Zeroshot} & llama3-8b & 75.0 & 52.2 & 53.5 & 69.8 & 70.0 & 39.2 & 35.5 & 21.6 & 75.6 \\
 & llama3-70b & 76.5 & 57.4 & 54.5 & 79.4 & 74.3 & 55.0 & 32.8 & 42.1 & 63.5 \\
 & chat-gpt & 60.0 & 63.2 & 49.0 & 60.0 & 71.0 & 57.0 & 41.0 & 37.0 & 69.0 \\
 & gpt-4o & 75.7 & 64.9 & 62.2 & 82.9 & 84.0 & \textbf{64.9} & \textbf{53.2} & 26.2 & 81.1 \\ \hline
\multirow{3}{*}{Fewshot} & llama3-8b & 74.9 & 53.2 & 43.9 & 73.3 & 74.8 & 49.0 & 18.5 & 43.0 & 70.1 \\
 & llama3-70b & 79.2 & 63.0 & 62.2 & 82.8 & 78.9 & 61.5 & 32.8 & 53.9 & 69.1 \\
 & chat-gpt & 71.0 & 64.0 & 59.0 & 84.0 & 83.0 & 52.0 & 27.0 & 48.0 & 72.0 \\
 & gpt-4o & 83.3 & 67.9 & 63.4 & 85.7 & 81.7 & 61.1 & 41.9 & \textbf{64.8} & 76.9
\\ \bottomrule

% existing/blackbox
\multirow{3}{*}{\makecell[l]{Out of the \\ box Systems}}
 & llama-g & 55.0  & - & 16.1 &  - & 75.5 &  -  & 43.6 & -  & 68.2 \\
 & openai-m & 72.0  & - & 63.1 & - & 73.0 & - & 46.3  &  -  & 75.9 \\
 & Perspective & 70.0  & - & \textbf{64.0}  & 89.0  &  81.0 & - & -  &   53.0 & 44.0
\\ \bottomrule

\end{tabular}
}
\caption{%Macro F1 scores for the binary and multi-label settings for the fine-tuned and zero-/few-shot models tested. The F1 scores for each class in the multi-label setting are also reported.
\dimos{Macro F1 scores for fine-tuned and zero-/few-shot models are reported in both binary and multi-label settings. Additionally, the F1 scores for each class in the multi-label setting are provided. For out-of-the-box systems, we report the F1 scores in the binary setting and, when available, the F1 scores achieved in each class.}}
\label{tab:results}
\end{table*}

\subsubsection{Out of the box Systems}
\label{sec:out-of-the-box}

The need for detecting sensitive or harmful content has led to several companies to develop their own models, which are made publicly available. In order to highlight the relevance of existing models for this task, we selected three popular specialised systems.

\noindent\textbf{Google's Perspective API (Perspective)} \cite{google_perspective_api} is a tool developed to detect and score various attributes of text, such as toxicity, and acts as a baseline performance of a production-ready API. In total Perspective provides scores for 16 different categories but in our use case we focus only on 12 of them that fit our taxonomy better\footnote{\label{note1}Detailed taxonomy can be found in Appendix \ref{sec:appendix-models}}.

\noindent \textbf{OpenAI's moderation API (openai-m)} is an endpoint tailored for content moderation. It classifies content into 18 potentially sensitive categories, 15 of which we map to our own taxonomy.
% \noindent\textbf{embed-english-v2.0 (cohere)} from Cohere\footnote{https://cohere.com/} which is designed to produce high quality embeddings of text to be used for tasks such as classification and semantinc similarity. For our use case the Classify endpoint of Cohere's API was utilised.

\noindent\textbf{Meta-Llama-Guard-2-8B (llama-g)} \cite{inan2023llama} is a specialised version of LLama-3 that aims to classify content based on a safety risk taxonomy of 11 harm categories \cite{vidgen2024introducing}\footnoteref{note1}. For our use case we consider only 5 of the categories that correspond better to the taxonomy used in {\DATASET}. Specifically we consider: "Hate" for \textit{Conflictual}, "Suicide \& Self-Harm" for \textit{self-harm}; and "Sexual Content", "Sex-Related Crimes", and "Child Sexual Exploitation" for \textit{sexual explicit content}.
 
% \begin{figure}
%     \centering
%     \includegraphics[width=1\linewidth]{images/out_of_the_box.png}
%     \caption{\fr{TODO: delete and merge lines in table 3}F1 scores for out of the box Systems. The macro-F1 scores for the binary and multi-label settings are displayed along with the F1 scores achieved in each class when it is available to the system tested.}
%     \label{fig:box_results}
% \end{figure}

\subsection{Evaluation Metrics}
Given the critical nature of the task and the importance of accurately identifying and recalling all potentially harmful content, we utilise F1 scores to evaluate our models. We assign equal weight to each label and report the macro-F1 score in both binary and multi-label settings. The F1 scores for individual labels are also considered in the multi-label scenario. This approach helps us gain a deeper insight into the challenges posed by the dataset.

\section{Results}
The scores for both binary and multi-label scenarios, across all models tested in the fine-tuning and zero-/few-shot settings, are presented in Table \ref{tab:results}. In general, fine-tuning leads to clear improvement for all models, which reinforces the importance of our dataset not only to evaluate models, but to build specialised models based based on it. \textit{llama3-8b} performs best overall, with macro-f1 scores of 85.6 in the binary setting, and 69.8 in the more challenging and fine-grained multi-label setting. %With the exception of \textit{chat-gpt} in the zero-shot setting, the majority of models perform significantly better in the binary setting compared to the multi-label one, exhibiting an average 14.5\% increase in macro-F1 score. This performance difference is expected when considering the fine-grained nature of the multi-label task and the imbalanced characteristics of our dataset.

\subsection{Fine-tuned Systems}
All the fine-tuned models demonstrate high performance  with \textit{RoBERTa} as the least effective, achieving macro-F1 scores of 82.4 in the binary setting and 64.7 in the multi-label setting. The specialised training corpus of \textit{tlm} appears to enhance its performance, as it consistently surpasses \textit{RoBERTa} of the same architecture in both settings. Moreover, the fine-tuned version of the larger and more recent \textit{llama3-8b} model achieves the best overall results in both settings, with macro-F1 scores of 85.6 in the binary setting and 69.8 in the multi-label setting, notably achieved without any hyper-parameter tuning, unlike the other models. Overall, the fine-tuned models tend to struggle the most with the least represented classes, such as \textit{Drugs}, \textit{Self-harm}, and \textit{Spam}. Interestingly, despite comprising 17.3\% of the total entries, the models under perform in the \textit{Conflictual} category, while they exhibit better performance in the less prevalent \textit{Sexual Explicit Content} class, which accounts for only 9.7\% of tweets. This disparity may indicate the models' difficulties in identifying subtler features within the \textit{Conflictual} category.

\subsection{Zero/Few-shot}

\paragraph{Zero-shot.}When tested without any contextual information, the models display varying degrees of effectiveness. Notably, the 70b version of llama3, \textit{llama3-70b}, outperforms its smaller counterpart and competes with OpenAI's models, achieving the highest macro F1 score of 76.5 in the binary setting. In the multi-label scenario, the \textit{gpt-4o} model excels, achieving a macro-F1 score of 64.9. Generally, the zero-shot models do not reach the performance levels of their fine-tuned counterparts, with the notable exception of \textit{gpt-4o}, which surpasses only \textit{RoBERTa} in the multi-label setting.

\paragraph{Few-shot.}
In the few-shot setting, \textit{llama3-8b} exhibits performance comparable to its zero-shot execution, illustrating the constraints of smaller models. This limitation is further highlighted by the performance gains observed in the rest of the models, \textit{llama3-70b}, \textit{chat-gpt}, and \textit{gpt-4o}, which show average increases of 7.6 and 3.2 points in macro-F1 for the binary and multi-label settings, respectively. This underscores the effectiveness of in-context learning in larger models. Overall, \textit{gpt-4o} achieves the best performance, competing with the fine-tuned models and notably outperforming \textit{llama3-8b} in specific categories,  \textit{Conflictual}, \textit{Drugs}, and \textit{Spam}. 

\subsection{Out of the box}
%Figure \ref{fig:box_results} presents the F1 scores of the "out of the box" systems evaluated (Section \ref{sec:out-of-the-box}). Generally, these models do not achieve high scores, except in specific categories like \textit{Profanity} and \textit{Sexual Explicit Content} where the \textit{Perspective} and \textit{LLama-g} systems excel. The \textit{Perspective} API manages to reach a macro-F1 of 70 in the binary setting, which still is the lowest overall, except for \textit{chat-gpt} in the zero-shot scenario. While these scores reflect the capabilities of these systems, it is important to note that, apart from \textit{Cohere} which was provided examples for each category, the \textit{Perspective} and \textit{Llama-g} do not use the same taxonomy as {\DATASET}, which can impact their performance.
 \dimos{When evaluating the performance of "out-of-the-box" models, we find that they generally fail to achieve high scores (Table \ref{tab:results}). Notable exceptions occur in specific categories such as \textit{Profanity} and \textit{Sexual Explicit Content}, where the \textit{Perspective} and \textit{llama-g} systems excel. The best performing out-of-the-box model, \textit{openai-m} API, achieves a macro-F1 score of 72\% in the binary setting, demonstrating greater robustness in detecting non-sensitive content (F1: 75.9\%) compared to its peers. Despite this, its overall performance remains the lowest among the models tested, except for \textit{chat-gpt} in zero- and few-shot scenarios. It is important to note that these scores may be influenced by the fact that these systems do not utilise the same taxonomy as {\DATASET}, which can impact their performance.}

% \begin{table}[]
% \scalebox{0.85}{
% \begin{tabular}{lccc}
% \toprule
%  & \textbf{Cohere} & \textbf{Perspective} & \textbf{Llama-guard} \\ \toprule
% \textbf{binary} & 50.95 & 70 & 40.68 \\ 
% \textbf{multi-label} & 38.48 & - & - \\ \midrule
% \textbf{conflictual} & 49.94 & 64 & 16.08 \\ 
% \textbf{profaninty} & 26.59 & 89 & - \\ 
% \textbf{sex} & 58.78 & 81 & 74.29 \\ 
% \textbf{drugs} & 44.01 & - & - \\ 
% \textbf{selfharm} & 17.89 & - & - \\ 
% \textbf{spam} & 39.86 & 53 & - \\ 
% \textbf{not sensitive} & 32.26 & 44 & 67.72 \\
% \bottomrule
% \end{tabular}
% }
% \caption{F1 scores for out of the box Systems. \dimos{Similar With figure fig:box\_results}}
% \label{tab:box_results}
% \end{table}

\section{Error Analysis}
Aiming to understand better the dataset and the challenges that the models face on identifying sensitive content we consider the best performing model, the fine-tuned \textit{llama3-8b}, and try to understand better its performance.

\begin{figure}
    \centering
    \includegraphics[width=1\linewidth]{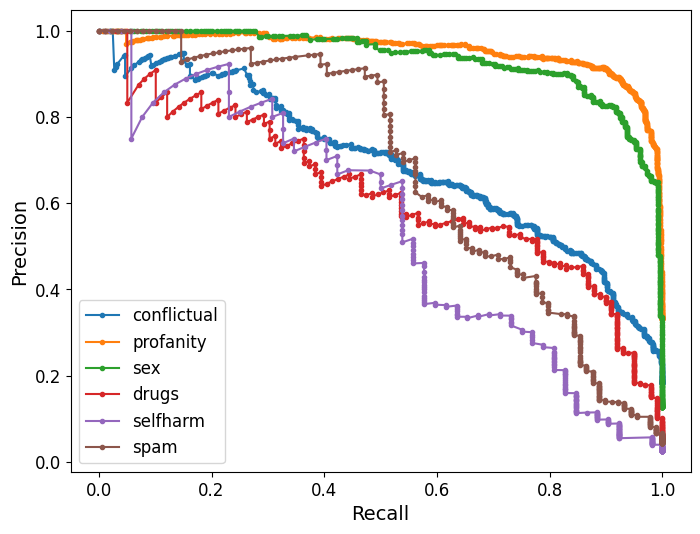}
    \caption{Precision-Recall curve for the fine-tuned \textit{llama3-8b} in the multi-label setting.}
    \label{fig:prec-recall_multi}
\end{figure}

In the binary setting the model displays a strong performance and achieves high precision and recall values, 85.5\%, 85.7\% respectively, signifying its ability to effectively identify true positive cases with a relatively low number of false positives (Figure \ref{fig:prec-recall_bin}, Appendix \ref{sec:appendix-results}). In contrast, for the multi-label  setting, the model seems to struggle with several categories as seen in Figure \ref{fig:prec-recall_multi}. 
%Despite \textit{llama-8b} achieving strong performance for the Profanity and Sex labels, the Conflictual, Drugs, self-harm, and Spam categories present a challenge for the classifier. Their precision drops rapidly as recall increases. 
%indicating that while the classifier captures more true positives, it also generates a higher number of false positives. 
%This same behaviour is observed for the Self-harm category, which exhibits the worst performance.
\fr{Despite the strong performance of \textit{llama-8b} in classifying Profanity and Sex labels, the model struggles with the Conflictual, Drugs, Self-harm, and Spam categories. As recall increases, precision for these categories drops significantly. This poses a particular challenge for health-related categories like Drugs and Self-harm, where high recall is critical, as missing cases could have serious consequences.}

% \dimos{TODO keywords check}

\section{Conclusions}

In this paper, we presented a complete research approach into sensitive content moderation in social media. Going beyond hate speech, we focus on categories that need to consistently be monitored in social media, let it be to filter to adult users or to remove from the platform, among others. We construct a multi-label dataset using six categories. The results show that LMs fine-tuned on our datasets are generally robust, although there are some categories where they are less precise, and hence these models are probably to be used as a support for human moderators. Nonetheless, the fact that these models perform at a high accuracy represents a useful tool to filter the most relevant messages for each category.

\section{Limitations}
\label{sec:limitations}
In this paper, we introduce a valuable new resource expected to benefit a wide range of researchers and industry professionals. However, it is important to acknowledge several limitations. Firstly, the dataset is limited in size, which may restrict the generalisability and robustness of the models trained on it. Additionally, it exclusively contains English-language content due to budget constraints, potentially overlooking the nuances and challenges present in other languages.

The methodology used for aggregating the data in our dataset (Section \ref{sec:annotation}) may also be subject to differing opinions. To facilitate transparency and further research, we plan to release all the collected annotations along with the dataset version used in our experiments. Moreover, the dataset was curated based on a specific selection of keywords, which might introduce biases and limit the diversity of the content. Another limitation is that the dataset is derived from only one social media platform, which may not fully represent the variety of sensitive content found across different platforms and contexts.

Finally, while we conduct an in-depth analysis using the results of six different models, there is significant room for improvement in terms of analysis and model development. This includes, investigating the performance of models of different architectures and optimising the prompts used\footnote{The prompts used in our experiments can be found in Appendix \ref{sec:appendix-prompts}}. %, and potentially incorporating the highlights collected in the classification process.

\section{Ethics Statement}
\label{sec:ethics}
We recognize the significance of the ACL Code of Ethics and are dedicated to adhering to its guidelines in our proposed task. Since our task involves user-generated content, we ensure user privacy by replacing each user mention in the texts with a placeholder, recognising the importance of anonymity, especially taking into account the potential for harm towards people expressing self-harming tendencies.

We also ensure fair treatment of the annotators who labelled the dataset by: 1) compensating them fairly at an average rate of 12\$ per hour, and 2) not sharing or storing personal identification information. As annotator demographics play an important role in the perception of toxicity, following~\citet{prabhakaran2021releasing}, we release the data\footnote{\url{https://huggingface.co/datasets/cardiffnlp/x_sensitive/blob/main/all_annotations.json}}, disaggregated by individual annotator labels, while making sure that the demographic information is course enough to prevent deanonymization of the crowd-workers. 

Lastly, recognise the sensitive and potentially dangerous nature of the dataset. However, we believe it is crucial to address and combat such behaviours. {\DATASET} will be shared under the CC BY-NC 4.0 Deed (Attribution-NonCommercial 4.0 International) following best practices in sharing social media-based data collections~\cite{fiesler2018participant,assenmacher2020end}. 

\section{Acknowledgments}
A big thank you to Debora Nozza for helpful discussions and feedback on the initial part of this project. Dimosthenis Antypas and Jose Camacho-Collados are supported by a UKRI Future Leaders Fellowship.

\bibliography{anthology,custom}
\appendix

\section{Data Collection}
\label{sec:appendix-data}

Table \ref{tab:data-sources} displays the sources used for finding keywords used to collect tweets.

\begin{table}[h]
\scalebox{0.7}{
\begin{tabular}{l|ll}
\textbf{Category} & \textbf{Sources} & \textbf{} \\ \hline
\multirow{5}{*}{Self-harm} & \cite{chancellor2016quantifying} &  \\
 & \cite{chancellor2016thyghgapp} &  \\
 & \cite{chancellor2016post} &  \\
 & \cite{chancellor2016recovery} &  \\
 & Ngrams from the Reddit training corpus &  \\ \hline
\multirow{3}{*}{\begin{tabular}[c]{@{}l@{}}Conflictual, \\ profane,\\ sexually explicit\end{tabular}} & \href{https://github.com/IDEA-NTHU-Taiwan/porn_ngram_filter}{IDEA-NTHU-Taiwan} &  \\
 & \href{https://github.com/LDNOOBW/List-of-Dirty-Naughty-Obscene-and-Otherwise-Bad-Words}{LDNOOBW} &  \\
 & \href{https://github.com/facebookresearch/flores/tree/main/toxicity} &  \\ \hline
Drugs & \href{https://www.talktofrank.com/drugs-a-z}{talktofrank} &  \\ \hline
% Guns & Wikipedia &  \\ \hline
\multirow{2}{*}{Spam} & \cite{founta2018large} &  \\
 & \#f4f, \#l4l, \#follow4follow, \#like4like & 
\end{tabular}}
\caption{Sources for seed words used for expanding our keyword lists.}
\label{tab:data-sources}
\end{table}

Table \ref{tab:seeds-stats} displays the total number of keywords and seed words used in our data collection.

\begin{table}[h]
\begin{tabular}{l|ll}
\textbf{Category} & \textbf{\#seed-words} & \textbf{\#keywords} \\ \hline
\begin{tabular}[c]{@{}l@{}}Conflictual, \\ profane,\\ sexually explicit\end{tabular} & 203 & 1322 \\ \hline
spam & 76 & 767 \\ \hline
drugs & 118 & 519 \\ \hline
self-harm & 56 & 734
\end{tabular}
\caption{Overall summary of the keyword list.}
\label{tab:seeds-stats}
\end{table}

% Figure \ref{fig:word_shift} displays word shift graphs  for conflictual, profane and sexual expclicit content. Many keywords are common between the three categories, particularly swear words and gendered slurs. 
% \begin{figure*}[h]
%     \centering
%     \includegraphics[width=1\linewidth]{images/word_shift.png}
%     \caption{word shift graphs for conflictual content, profanity, and sexually explicit content. 
% }
%     \label{fig:word_shift}
% \end{figure*}

\section{Annotation}
\label{sec:appendix_annotation}
\subsection{Guidelines}

\begin{figure}[h]
    \centering
    \includegraphics[width=1\linewidth]{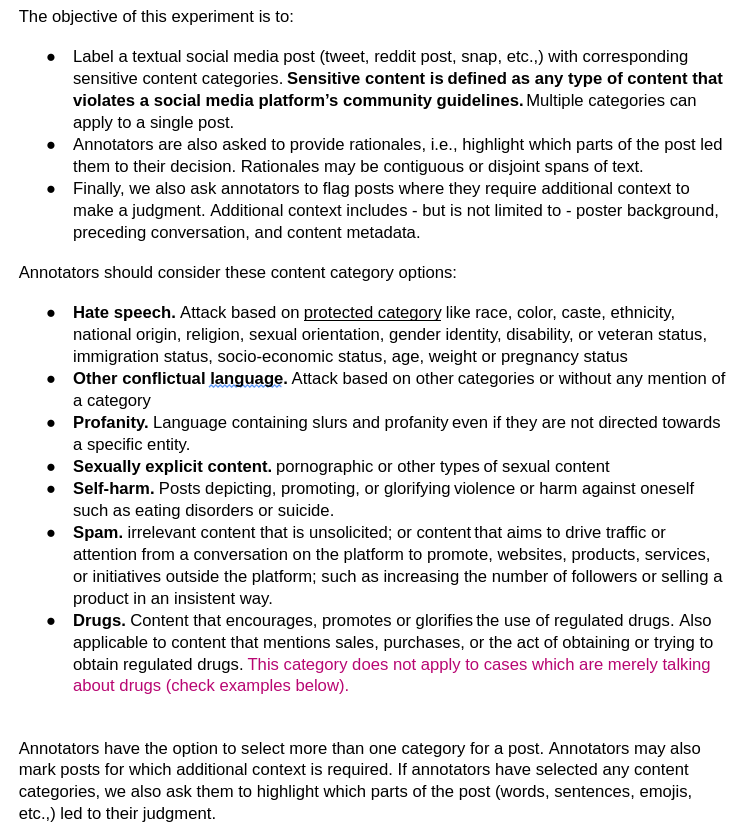}
    \caption{Guidelines provided to annotators.}
    \label{fig:appendix_guidelines}
\end{figure}

Figure \ref{fig:appendix_guidelines} displays the guidelines provide to each coder for the annotation task.

\subsection{Annotator Agreement}
Table \ref{tab:agreement_gender} displays the inter annotator agreement based on Krippendorff's alpha between different gender groups of coders. Similarly, Table \ref{tab:agreement_age} displays agreement scores between different codes age groups.

\begin{table}[hb]
\centering
\begin{tabular}{lcc}
Gender & Multi & Bin \\ \hline
Man & 0.49 & 0.57 \\
Woman & 0.49 & 0.55 \\
Non-binary & 0.47 & 0.82 \\
\end{tabular}
\caption{Krippendorff's alpha within each gender group of coders.}
\label{tab:agreement_gender}
\end{table}

\begin{table}[]
\centering
\begin{tabular}{lcc}
Age & Multi & Bin \\ \hline
18-25 & 0.51 & 0.57 \\
26-39 & 0.51 & 0.59 \\
40-65 & 0.47 & 0.52 \\
over 65 & 0.47 & 0.54
\end{tabular}
\caption{Krippendorff's alpha within each age group of coders.}
\label{tab:agreement_age}
\end{table}

Tables \ref{tab:coder_age} and \ref{tab:coder_gender} show the percentage of tweets labelled as each class by age and gender groups of coders, respectively.

\begin{table}[]
\scalebox{0.55}{
\begin{tabular}{lcccccc|c|c}
\textbf{Age} & \textbf{conflictual} & \textbf{profanity} & \textbf{sex} & \textbf{drugs} & \textbf{self-harm} & \textbf{spam} & \textbf{AVG} & \textbf{Coders} \\ \hline
18-25 & 15 & 28 & 9 & 3 & 3 & 4 & 10 & 77 \\
26-39 & 13 & 28 & 9 & 3 & 3 & 4 & 10 & 269 \\
40-65 & 15 & 25 & 8 & 3 & 2 & 3 & 9 & 166 \\
over 65 & 10 & 27 & 5 & 3 & 1 & 2 & 8 & 10
\end{tabular}}
\caption{Percentage of tweets labelled as each class for each age bracket of coders. }
\label{tab:coder_age}
\end{table}

\begin{table}[h]
\scalebox{0.55}{
\begin{tabular}{l|cccccc|c|c}
\textbf{Gender} & \textbf{conflictual} & \textbf{profanity} & \textbf{sex} & \textbf{drugs} & \textbf{self-harm} & \textbf{spam} & \textbf{AVG} & \textbf{Coders} \\ \hline
Woman & 14 & 27 & 8 & 3 & 3 & 3 & 10 & 268 \\
Non-Binary & 20 & 28 & 10 & 2 & 3 & 3 & 11 & 12 \\
Man & 14 & 26 & 9 & 3 & 3 & 4 & 10 & 241
\end{tabular}
}
\caption{Percentage of tweets labelled as each class for each gender of coders. }
\label{tab:coder_gender}
\end{table}

\section{Models}
\label{sec:appendix-models}
\subsection{Resources}
In total we estimate 112 hours used for the training of  \textit{RoBERTa, tlm} and  \textit{llama3-8b} models using a NVIDIA GeForce RTX 4090  GPU and 90 hours for inferences with the \textit{llama3-8b} and \textit{llama3-70b}  models using an  NVIDIA Quadro RTX 8000 GPU.
Table \ref{tab:parameters} provides details for the models used in our experiments.

\begin{table}[ht]
\centering
\begin{tabular}{l|c}

\textbf{Model}         & \textbf{Parameters} \\ \hline
RoBERTa                             & 355M                  \\ \hline
tlm                           & 355M                  \\ \hline
  llama3-8b                         & 8B  \\ \hline
llama3-70b                         & 70B  \\ \hline
chat-gpt                      &  175B (approximate)     \\ 
\end{tabular}
\caption{Number of Parameters in different language models used.}
\label{tab:parameters}
\end{table}

\subsection{Taxonomies}
\paragraph{Taxonomy used by Google's perspective API:}
\begin{enumerate}
  \item TOXICITY
  \item SEVERE\_TOXICITY
  \item IDENTITY\_ATTACK
  \item INSULT
  \item PROFANITY
  \item SEXUALLY\_EXPLICIT
  \item THREAT
  \item FLIRTATION
  \item ATTACK\_ON\_AUTHOR
  \item ATTACK\_ON\_COMMENTER
  \item INCOHERENT
  \item INFLAMMATORY
  \item LIKELY\_TO\_REJECT
  \item OBSCENE
  \item SPAM
  \item UNSUBSTANTIAL
\end{enumerate}
In our experiments we utilise the following class mapping to the {\DATASET} taxonomy: "TOXICITY": Conflictual,  "PROFANITY": Profanity, "SEXUALLY\_EXPLICIT": Sexual Explicit Content,
"SPAM": Spam.

\paragraph{MLCommons taxonomy used in Meta-LLama-guard:}
\begin{itemize}
    \item[1:] Violent Crimes
    \item[2:] Non-Violent Crimes
    \item[3:] Sex-Related Crimes
    \item[4:] Child Sexual Exploitation
    \item[5:] Specialized Advice
    \item[6:] Privacy
    \item[7:] Intellectual Property
    \item[8:] Indiscriminate Weapons
    \item[9:] Hate
    \item[10:] Suicide \& Self-Harm
    \item[11:] Sexual Content
\end{itemize}

\paragraph{Openai's moderation API taxonomy:}
\begin{itemize} \item[1:] harassment \item[2:] harassment\_threatening \item[3:] hate \item[4:] hate\_threatening \item[5:] self\_harm \item[6:] self\_harm\_instructions \item[7:] self\_harm\_intent \item[8:] sexual \item[9:] sexual\_minors \item[10:] violence \item[11:] violence\_graphic \item[12:] self-harm \item[13:] sexual/minors \item[14:] hate/threatening \item[15:] violence/graphic \item[16:] self-harm/intent \item[17:] self-harm/instructions \item[18:] harassment/threatening \end{itemize}
In our experiments we utilise the following class mapping to the {\DATASET} taxonomy: 
\begin{verbatim}
{
'harassment': 'Conflictual',
'harassment_threatening': 'Conflictual',
'hate': 'Conflictual',
'hate_threatening': 'Conflictual',
'self_harm': 'Self-Harm',
'self_harm_instructions': 'Self-Harm',
'self_harm_intent': 'Self-Harm',
'sexual': 'Sexual Explicit Content',
'sexual_minors': 'Sexual Explicit Content',
'self-harm': 'Self-Harm',
'sexual/minors': 'Sexual Explicit Content',
'hate/threatening': 'Conflictual',
'self-harm/intent': 'Self-Harm',
'self-harm/instructions': 'Self-Harm',
'harassment/threatening': 'Conflictual'
}
\end{verbatim}

\subsection{Prompts}
\label{sec:appendix-prompts}
Below are the prompt templates used in the zero- and few-shot settings.

\noindent\textbf{Multi-label}

\noindent Classify the following sentence in one or more of the categories listed below: 1. self-harm, 2. drugs, 3. conflictual language, 4. spam, 5. profane or bad words, 6 sexually explicit. Answer with the following format: 'ANSWER= 2, 3' if the sentence contains sensitive content or 'ANSWER= 0' if the sentence does not contain sensitive content. 

\noindent Sentence: "\{\{ message \}\}"

\noindent\textbf{Binary}

\noindent Below there is a sentence. I would like to know if it contains sensitive content. We understand the sentence contains sensitive content if it treats topics like self-harm, drugs, if it contains conflictual language, spam, expresses profanity or it is sexually explicit. Does this tweet contain any kind of sensitive content? Please answer with the following format: 'ANSWER= 1' if the tweet contains sensitive content or 'ANSWER= 0' if the sentence does not contain sensitive content.

\noindent Sentence: "\{\{ message \}\}"

\subsection{Extended Results}
\label{sec:appendix-results}
\begin{figure}
    \centering
    \includegraphics[width=1\linewidth]{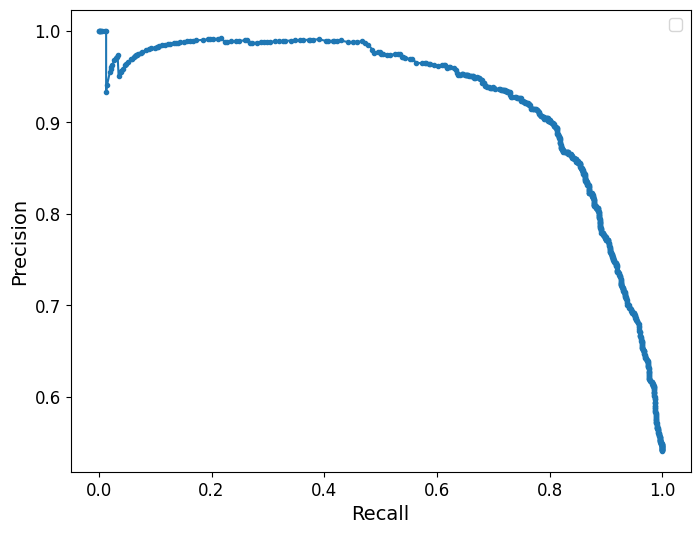}
    \caption{Precision-Recall curve for the fine-tuned \textit{llama3-8b} model in the binary setting. }
    \label{fig:prec-recall_bin}
\end{figure}

Figure \ref{fig:prec-recall_bin} displays the Precision-Recall curve for the fine-tuned \textit{llama3-8b} model in the binary setting.

\end{document}